\documentclass[runningheads]{llncs}
\usepackage{graphicx}

\usepackage{tikz}
\usepackage{comment}
\usepackage{amsmath,amssymb} 
\usepackage{color}

\usepackage[accsupp]{axessibility}  

\begin{document}
\pagestyle{headings}
\mainmatter
\def\ECCVSubNumber{30}  

\title{CEN-HDR: Computationally Efficient neural Network for real-time High Dynamic Range imaging} 

\titlerunning{CEN-HDR}
%
\author{Steven Tel\orcidID{0000-0002-1487-9381} \and
Barthélémy Heyrman\orcidID{0000-0003-1642-8311} \and
Dominique Ginhac\orcidID{0000-0002-5911-2010}}
\authorrunning{S. Tel et al.}
%
\institute{ImViA EA7535, University Burgundy Franche-Comté, 21078 Dijon, France 
\email{\{steven.tel,barthelemy.heyrman,dginhac\}@u-bourgogne.fr}}
\maketitle

\begin{abstract}
High dynamic range (HDR) imaging is still a challenging task in modern digital photography. Recent research proposes solutions that provide high-quality acquisition but at the cost of a very large number of operations and a slow inference time that prevent the implementation of these solutions on lightweight real-time systems.
In this paper, we propose CEN-HDR, a new computationally efficient neural network by providing a novel architecture based on a light attention mechanism and sub-pixel convolution operations for real-time HDR imaging. We also provide an efficient training scheme by applying network compression using knowledge distillation. We performed extensive qualitative and quantitative comparisons to show that our approach produces competitive results in image quality while being faster than state-of-the-art solutions, allowing it to be practically deployed under real-time constraints. Experimental results show our method obtains a score of 43.04 $\mu$-PSNR on the \textit{Kalantari2017} dataset with a framerate of 33 FPS using a Macbook M1 NPU. The proposed network will be available at \textit{https://github.com/steven-tel/CEN-HDR}
\end{abstract}

\section{Introduction}

\par In the last decades, applications based on computer vision have become increasingly important in everyday life. Currently, many research works are conducted to propose more reliable algorithms in areas such as object detection, action recognition, or scene understanding. However, the accuracy of these algorithms depends largely on the quality of the acquired images. Most standard cameras are unable to faithfully reproduce the illuminations range of a natural scene, as the limitations of their sensors generate a loss of structural or textural information in under-exposed and over-exposed regions of the acquired scene. To tackle this challenge, sensors with a higher dynamic range (HDR) have been proposed \cite{855857,1467255} to capture more intensity levels of the scene illumination, but these solutions are expensive, preventing high dynamic range acquisition from being readily available. 

Towards making HDR imaging practical and accessible, software solutions were proposed, based on the emergence of deep learning in computer vision applications. They acquire one Low Dynamic Range (LDR) image and try to expends its dynamic range thanks to a generative adversarial network \cite{liu2020singleimage,9206076,khan2019fhdr}. Although these methods produce images with a higher illumination range, they have limitations in extending the dynamic of the input image to the dynamic of the acquired scene. A more effective approach is to acquire multiple LDR images with different exposure times and merge them into one final HDR image. Traditional computer vision algorithms \cite{devebec} allow the acquisition of good-quality static scenes when there is no camera or object motion between images with different exposure times. However, in a lot of use cases, images are captured in a rapid sequence from a hand-held device resulting in inevitable misalignments between low dynamic range shots. Therefore scenes with motions introduce new challenges as ghost-like artifacts for large motion regions or loss of details in occluded regions. Following recent advances in the field of deep learning, several methods based on Convolutional Neural Network (CNN) were proposed to spatially align input frames to a reference one before merging them into a final HDR image. 

State-of-the-art deep learning solutions for multi-frame merging HDR \cite{Qian,liu2021adnet} tend to be based on a previously proposed method\cite{ahdrnet} and add additional processing to increase the accuracy of the HDR merging system. As a consequence, the computational cost and execution time are significantly increased, preventing these solutions to be used in lightweight systems and/or in real-time applications. Then, the primary goal of HDR imaging software solutions, which was to make HDR imaging more widely available compared to hardware solutions, is therefore not being achieved at all. Therefore, in this paper, we propose a Computationally Efficient neural Network for High Dynamic Range imaging (CEN-HDR). CEN-HDR is based on an encoder-decoder neural network architecture for generating ghost-free HDR images from scenes with large foreground and camera movements. Unlike previously published solutions, we decided to develop a new approach keeping in mind the constraint of inference time and computational cost.

\begin{enumerate}
    \item We propose CEN-HDR a novel efficient convolutional neural network based on a new attention mechanism and sub-pixel convolution that overcomes ghost-like artifacts and occluded regions while keeping a low computational cost, allowing our solution to be implemented in real-time on a lightweight system.
    \item We demonstrate the efficiency of network compression for the realization of CEN-HDR by applying a knowledge distillation scheme.
    \item We perform extensive experiments to determine the best trade-off between accuracy and inference cost with the main objective to demonstrate the relevance of CEN-HDR.
\end{enumerate}

\section{Related Works}

We briefly summarize existing HDR merging approaches into two categories: deep learning-based architectures and efficient learning-based architectures. In the first category, the proposed methods aim to achieve better quality in HDR imaging without taking inference cost into account. Approaches belonging to the second category seek to optimize the compromise between the quality of the generated images and the computation cost. This leads to the use of new operators in the proposed deep learning architectures.

\begin{figure}[ht]
\centering
\includegraphics[height=4.2cm]{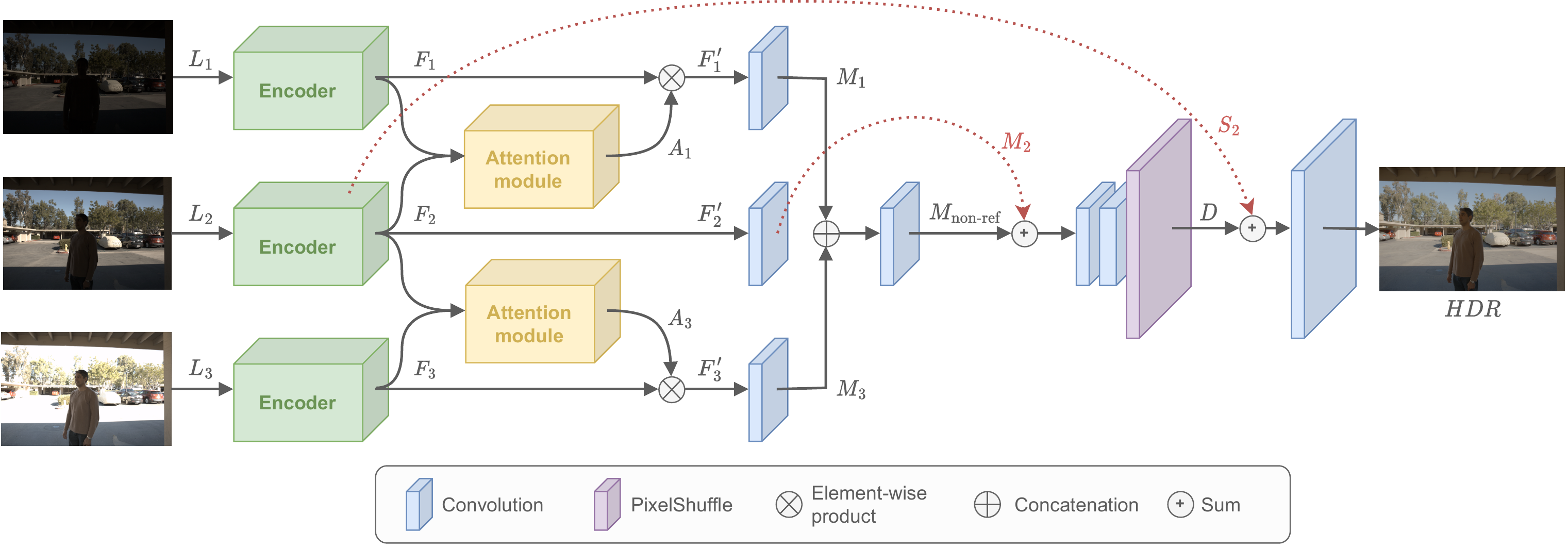}
\caption{Architecture of the proposed CEN-HDR solution. The spatial size of input features is divided by 2 at the encoding step. The attention module allows registering non-reference features to the reference ones. The full spatial size is recovered thanks to the pixel shuffle operation.}
\label{fig:network_arch}
\end{figure}

\subsection{Deep learning based HDR merging}
Using multiple input images for HDR generation leads to the need to align the features of the LDR images to the reference image.
The first common method for feature registration is by computing the motion between inputs features using an optical flow algorithm. Multiple studies \cite{8658831,Kalantari2017,rnn} used Liu\cite{celiu} optical flow algorithm. In \textit{Kalantari et al.}\cite{Kalantari2017}, the input images are aligned by selecting the image with better pixels as a reference and computing the optical flow between this reference and other input LDR images. Then, the warped images are fed to a supervised convolutional neural network (CNN) to merge them into an HDR image. However, since the optical flow algorithm initially assumes that the input images have the same exposure time, trying to warp the different exposures with occluded regions can result in artifacts in the final HDR image. To address this issue, \textit{DeepHDR}\cite{deephdr} proposes an image translation network able to hallucinate information in the occluded regions without the need for optical flow. Moreover, many solutions have been developed to correct the ghost effect introduced by the misalignment of the input images. In \textit{AHDRNet}\cite{ahdrnet} an attention module is proposed to emphasize the alignment of the input images to the reference image. Input images are then merged using several dilated residual dense blocks. The high performance of the AHDRNet network led it to be used as a base network for other methods. For example,  \textit{ADNet}\cite{liu2021adnet} follows the same main architecture that AHDRNet but adds a Pyramidal alignment module based on deformable convolution allowing a better representation of the extracted features but in the counterpart of a larger number of operations.

\subsection{Efficient learning-based HDR merging architectures}
To our best knowledge, the first architecture which aims to be efficient was proposed by  \textit{Prabhakar et al.}\cite{Prabhakar2020TowardsPA} by processing low-resolution images and upscaling the result to the original full resolution thanks to a bilateral guided upsampling module. 
Recently, the HDR community tends to focus more on efficient HDR image generation\cite{ntire22}  and no longer only aims at improving the image quality but also at significantly limiting the number of processing operations. This results in efficient solutions such as \textit{GSANet}\cite{Li_2022_CVPR} that propose efficient ways to process gamma projections of input images with spatial and channel attention blocks to increase image quality while limiting the number of parameters.
 \textit{Yu et al.}\cite{progresiveNTIRE} introduce a multi-frequency lightweight encoding module to extract features and a progressive dilated u-shape block for features merging. Moreover, the different standard convolution operations are replaced by depth-wise separable convolution operations firstly proposed in \cite{dsc}, they are composed of a depth-wise convolution followed by a pointwise convolution which allows more efficient use of model parameters. Another efficient method, proposed by \textit{Yan et al.}\cite{Yan_2022_CVPR} is a lightweight network based on an u-net\cite{unet} like encoder-decoder architecture, allowing for spatially reduced processed features. While these solutions focus on the number of performed operations, their inference time still remains too long to be considered as real-time solutions.

\section{Proposed Method}

We consider three LDR images $I_i \in \mathbb{R} ^{3 \times H \times W}$ with their respective exposure times $t_{i}$ as inputs. The generated HDR image is spatially aligned with the central LDR frame $I_{2}$ selected as the reference image. To make our solution more robust to exposure difference between inputs, the respective projection of each LDR input frame into the HDR domain is calculated using the gamma encoding function described in Eq.~\ref{eq:hi}, following previous works \cite{liu2021adnet,hdrgan,ahdrnet}:

. 
\begin{equation}\label{eq:hi}
H_{i} = \frac{I_{i}^\gamma}{t_{i}}, \quad \gamma=2,2
\end{equation}
Where $H_i \in \mathbb{R} ^{3 \times H \times W}$ is the gamma-projected input.
Then, each input is concatenated with their respective gamma-projection to obtain $L_i \in \mathbb{R} ^{6 \times H \times W}$ :
\begin{equation}\label{eq:li}
L_i = I_i  \oplus H_i
\end{equation}
where $\oplus$ represents the concatenation operation. $L_i$ will then be fed to our proposed merging network whose architecture is detailed in Fig.~\ref{fig:network_arch}.

\subsection{Feature encoding}
Using high-resolution images as inputs presents an additional challenge in the design of a real-time HDR merging network. To solve such a problem, previous works \cite{Wu_2018_ECCV,Yan_2022_CVPR} propose to use a U-net\cite{unet} like architecture to reduce the spatial size of the features processed by the merging network. However, a too large reduction of the spatial dimensions causes the extraction of coarse features that degrade the final result. So, we decide to limit the spatial reduction to 2 by using an encoder block composed of 2 sequential convolutions as described in Eq.~\ref{eq:fi}.
\begin{equation}\label{eq:fi}
F_i = conv_{E_1}(conv_{E_2}(L_i))
\end{equation}
where $F_i \in \mathbb{R} ^{32 \times \frac{H}{2} \times \frac{W}{2}}$ is the features map extracted from the encoder for each LDR input. $conv_{E_1}$ and $conv_{E_2}$ are 3x3-convolution layers extracting respectively 16 and 32 features map. The spatial size is divided by 2 setting a stride of 2 for $conv_2$.

\begin{figure}[ht]
\centering
\includegraphics[height=6cm]{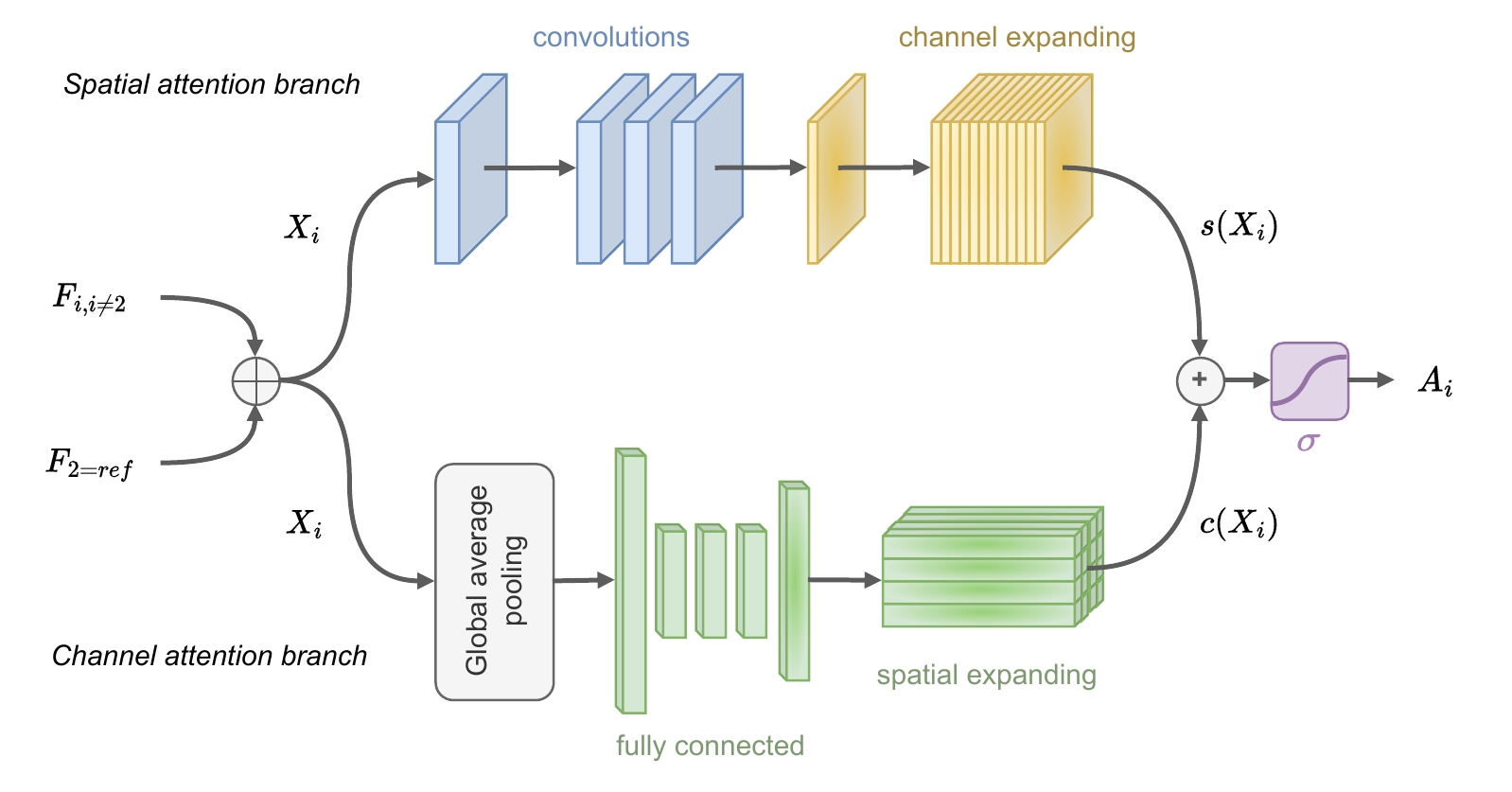}
\caption{Illustration of the attention module composed of 2 branches, respectively responsible for spatial attention and channel attention. A sigmoid activation function is used to keep the value between 0 and 1.}
\label{fig:attention_module}
\end{figure}

\subsection{Attention module}
The final generated HDR image must be aligned with the reference image. To address this requirement \cite{ahdrnet} demonstrates the effectiveness of using a spatial attention module after the encoding step. Since then, many spatial and channel attention modules have been proposed in the literature which can be integrated into networks to improve their performance. According to the inference cost study done in Table ~\ref{table:attentionmodulecmp}, we propose the \textit{Spatial-Channel Reference Attention Module} (SCRAM) a slightly modified version of the Bottleneck Attention Module (BAM) proposed in \cite{BAM}. Indeed while BAM aims to generate a mask of its input feature maps, in our case we want to generate attention maps from the concatenation of reference and non-reference features, these attention maps are then applied to non-reference features only resulting in a reduction of the number of feature maps in the proposed attention module. Moreover, batch normalization is not applied in SCRAM. The detailed structure of SCRAM is illustrated in Fig.~\ref{fig:attention_module}.

The non-reference features $F_{i\ne2}$ are concatenated with the features of the reference image:
\begin{equation}\label{eq:cat_attention}
X_{i} = F_{i} \oplus F_{2=ref}, \quad i\ne2
\end{equation}
where  $X_i \in \mathbb{R} ^{64 \times \frac{H}{2} \times \frac{W}{2}}$ is the input of SCRAM and $\oplus$ is the concatenation operator.

Following \cite{BAM}, SCRAM is composed of two branches respectively responsible for the spatial and channel features alignment of the non-reference images to the reference ones:

\begin{equation}\label{eq:outatt}
A_i = \sigma(s(X_i) + c(X_i)),  \quad i\ne2
\end{equation}
where $s$ is the spatial attention branch and $c$ is the channel attention branch. The sum of produced features by each branch passes through $\sigma$, a sigmoid activation function to keep the output values between 0 and 1. \\

\textit{Spatial attention:} The objective of the spatial branch is to produce an attention map allowing to keep the most relevant information for the spatial alignment of the non-reference images to the reference one. To limit the computation, we first reduce the number of features map by 3 using a pointwise convolution. With the objective to extract more global features  while keeping the same computation, we then make the receptive field larger by employing 3 dilated convolutions \cite{dilated_conv} layers with a factor of dilatation set to 2.  The final attention map of size $(1, H, W)$ is produced by using a pointwise convolution and then expanded across the channel dimension to obtain $A_S \in \mathbb{R} ^{32 \times H \times W}$.\\

\textit{Channel attention:}  This branch aims to perform a channel-wise feature recalibration. We first squeeze the spatial dimension by applying a global average pooling which sums out the spatial information to obtain the features vector of size $(64, 1, 1)$. A multilayer perceptron with three hidden layers is then used in purpose to estimate cross-channel attention. The last activation size is set to 32 to fit the number of channels of the non-reference features $F_i$. Finally, the resulting vector map is spatially expanded to obtain the final feature map $A_C \in \mathbb{R} ^{32 \times H \times W}$. \\

The Attention features $A_i$ are then used to weight the non-reference features $F_i$:

\begin{equation}\label{eq:ewmulatt}
F'_i = F_i \otimes A_i,  \quad i\ne2
\end{equation}
where $\otimes$ is the element-wise product and $F'_i$ is the aligned non-reference features. For the reference features we set $F'_2 = F_2$.

\subsection{Features merging}
While most of the computation is usually done in the merging block \cite{liu2021adnet,hdrgan,ahdrnet}, we propose a novel efficient feature merging block that first focuses on merging the non-references features. 
Each feature map $F'_{i}$ produced by the encoder goes through a convolution layer:
\begin{equation}
    M_i = conv_{M_{1}}(F'_{i})
\end{equation}
Where $conv_{M_{1}}$ is a 3x3-convolution producing $M_i \in \mathbb{R} ^{64 \times \frac{H}{2} \times \frac{W}{2}}$. 
Then we focus on merging the non-reference features maps by concatenating them and feeding the result features in a convolution layer:
\begin{equation}
    M_{\text{non-ref}} = conv_{M_{2}}(M_{1} \oplus M_{3})
\end{equation}
where $conv_{M_{2}}$ is a 3x3-convolution and $M_{\text{non-ref}} \in \mathbb{R} ^{64 \times \frac{H}{2} \times \frac{W}{2}}$ is the non-reference merged features. 

As we emphasize the reference features throughout all our network, here we merge our reference features with the non-reference $M_{non-ref}$  only by adding them together to limit the number of features map processed later:
\begin{equation}
    M = conv_{M_{4}}(conv_{M_{3}}(M_{2} + M_{non-ref}))
\end{equation}
where $conv_{M_{3}}$ and $conv_{M_{4}}$ are  3x3-convolutions producing each 64 features map and $M \in \mathbb{R} ^{64 \times \frac{H}{2} \times \frac{W}{2}}$ contains features from all LDR input images.

\begin{figure}[ht]
\centering
\includegraphics[height=4cm]{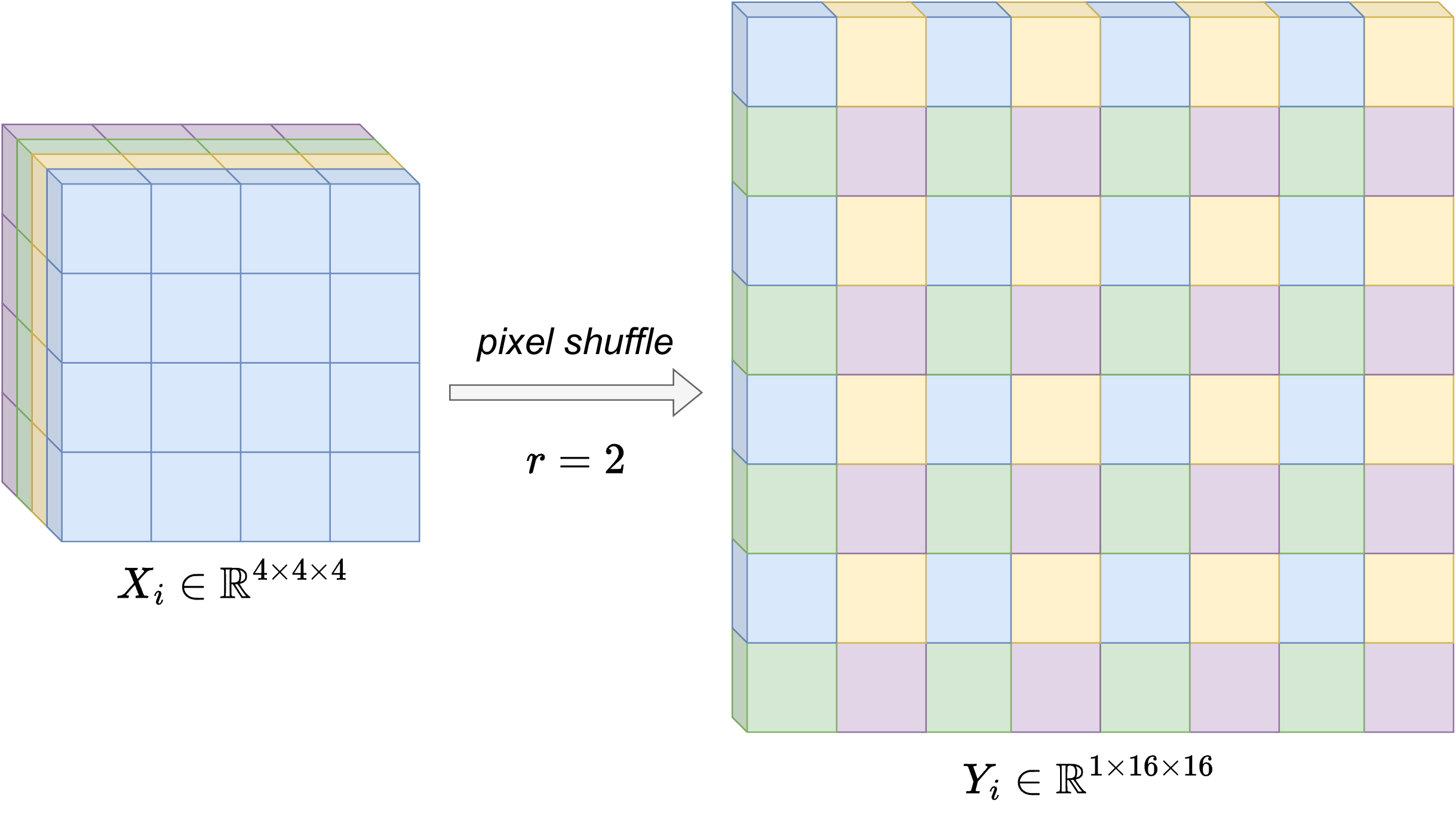}
\caption{Illustration of the pixel rearrangement by the pixel shuffle layer for an upscale factor set to $r=2$ and an input shape of $(4, 4, 4)$. In the proposed solution, the input shape is $(64, \frac{H}{2}, \frac{W}{2})$, the produced output size is $(16, H, W)$. }
\label{fig:pixelshuffle}
\end{figure}

\subsection{Features decoding}
The role of the decoder is to produce the final HDR image from the features produced by the merger block. At the encoding stage, we divided the spatial dimensions by 2. While the original spatial size is usually recovered using bilinear upsampling or transposed convolution\cite{Long_2015_CVPR} operation, we propose to use the pixel shuffle operation first proposed in \cite{shi2016realtime}, it is presented as an efficient sub-pixel convolution with a stride of $1/r$ where $r$ is the upscale factor. In our case, we set $r=2$. As illustrated in Fig.~\ref{fig:pixelshuffle}, the pixel shuffle layer rearranges elements in a tensor of shape $(C \times r^2, H, W)$ to a tensor of shape $(C, H\times r, W\times r)$:
\begin{equation}
    D = PixelShuffle(M)
\end{equation}
where $D \in \mathbb{R} ^{16 \times H \times W}$ is the resulting upscaled features.

The final HDR image is obtained by following the Eq \ref{eq:hdr}. 
\begin{equation}\label{eq:hdr}
    HDR = \sigma(conv_D(D + S_2))
\end{equation}
where $S_2$ is the reference features extracted by the first convolution layer of our network $conv_{E_1}$ to stabilize the training of our network. Finally, we generate the final HDR image using a 3x3-convolution layer, followed by a Sigmoid activation function. 

\section{Experimental Settings}

\subsection{Datasets}
The CEN-HDR network has been trained using the dataset provided by \cite{Kalantari2017} composed of 74 training samples and 15 test samples. Each sample represents the acquisition of a dynamic scene caused by large foreground or camera motions and is composed of three input LDR images (with EV of -2.00, 0.00, +2.00 or -3.00, 0.00, +3.00) and a reference HDR image aligned with the medium exposure image. The network has also been separately trained and tested using the dataset from the NTIRE\cite{ntire22} dataset, where the 3 LDR images are synthetically generated from the HDR images provided by \cite{Froehlich}. The dataset is composed of 1500 training samples, 60 validation samples, and  201 testing samples. The ground-truth images for the testing sample are not provided.

\begin{figure}[ht]
\centering
\includegraphics[height=6.3cm]{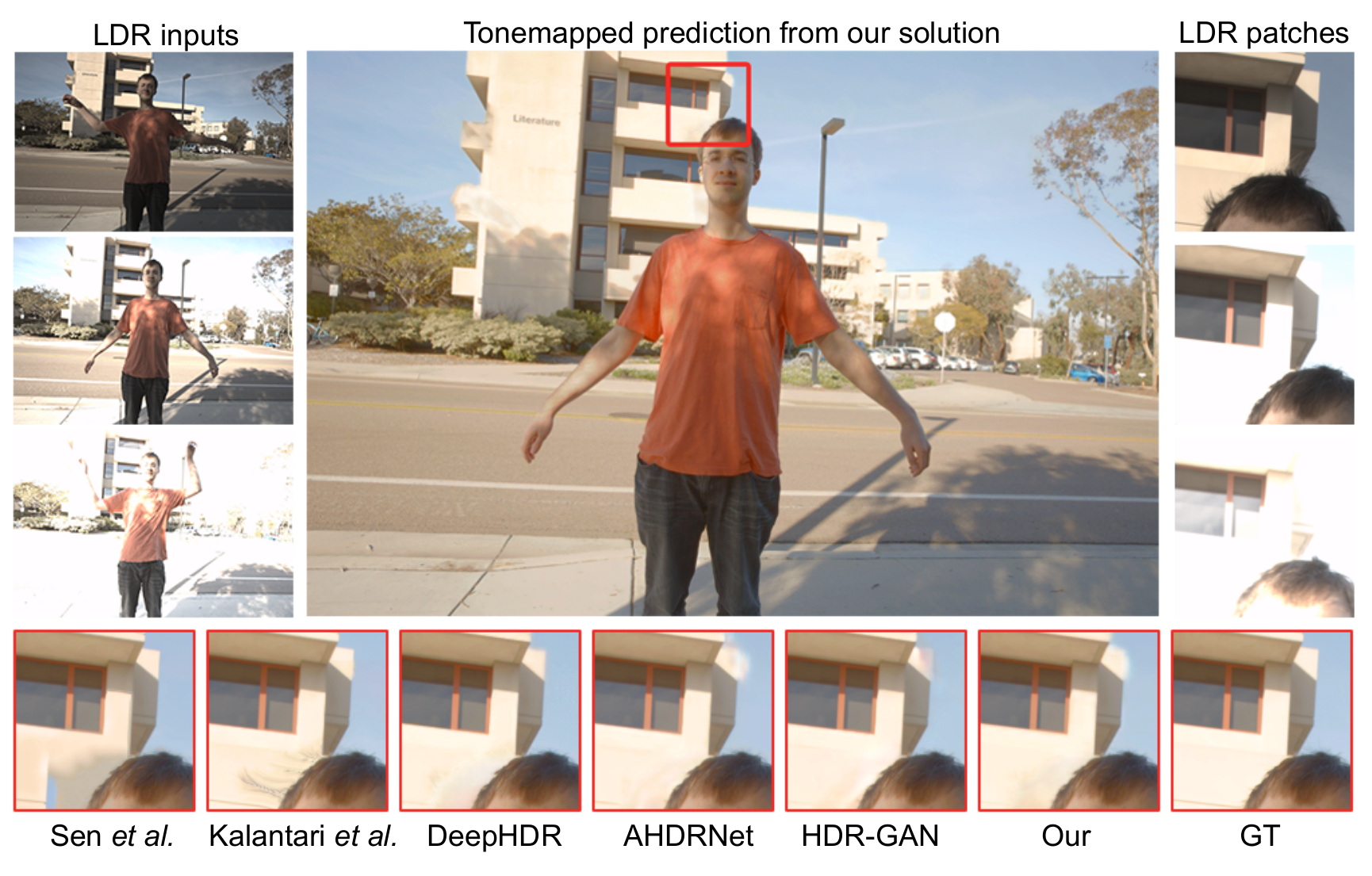}
\caption{Qualitative comparison of the proposed CEN-HDR solution with other HDR merging methods. The cropped patch demonstrates that the proposed efficient network has equivalent capabilities to the state-of-the-art methods to correct the ghost effect due to the large movement in the scene.}
\label{fig:qualitative_comparison}
\end{figure}

\subsection{Loss function}
Following previous works \cite{Kalantari2017,ahdrnet}, the images have been mapped from the linear HDR domain to the LDR domain before evaluating the loss function. In order to train the network, the tone-mapping function has to be differentiable around zero, so, the $\mu$-law function is defined as follows:

\begin{equation}\label{eq:mulaw}
T(H) = \frac{log(1+\mu H)}{log(1+\mu)}, \quad \mu=5000
\end{equation}
where H is the linear HDR image and $\mu$ the amount of compression.\\

To make an efficient network, a network compression method defined as knowledge distillation proposed in \cite{kd} has been used. By using knowledge distillation, we assume that the capacity of a large network is not fully exploited, so the objective is to transfer the knowledge of this large teacher network to our lighter network as described in the Eq.~\ref{eq:lossfunction}.

\begin{equation}
\label{eq:lossfunction}
\mathcal{L}=  \alpha \times \mathcal{L}(T, T_{GT}) + (1 - \alpha) \times \mathcal{L}(T, T_{Teacher})
\end{equation}
where $\mathcal{L}$ is the $L_1$ loss function. $T$ is the tone mapped prediction of our network, $T_{GT}$ the tone mapped ground truth provided in the dataset and $T_{Teacher}$ the tone mapped prediction of the large teacher network. we use the HDR-GAN \cite{hdrgan} model as the teacher. $\alpha$ is a trade-off parameter set to 0.2.
Moreover, the method proposed by \cite{Kalantari2017} to produce the training dataset focuses mainly on foreground motion. It does not allow the generation of reliable ground-truth images for chaotic motions in the background, such as the movement of tree leaves due to wind, which results in a ground truth image with blurred features that do not reflect reality. This has the effect of producing a greater error when the predicted image contains sharper features than in the ground truth image. Using also a predicted image from a teacher model allows for dealing with this data misalignment. The comparison of the performance obtained between training done with knowledge distillation and without is made in Table.~\ref{table:kd}. 

\subsection{Implementation details}

The CEN-HDR network has been trained using cropped patches of size $256 \times 256$ pixels with a stride of 128 and evaluated on the full-resolution test images. During training, random augmentations are applied on the cropped patch such as horizontal symmetry and rotation of 90, 180, or 270 degrees.
Training has been done using the Adam optimizer with a batch size of 8. The learning rate is initially set to $ 10^{-4}$, keep fixed for 80 epochs, and decreased by 0.8 every 20 epochs after. The training lasts for 500 epochs.

\begin{figure}[ht]
\centering
\includegraphics[height=6cm]{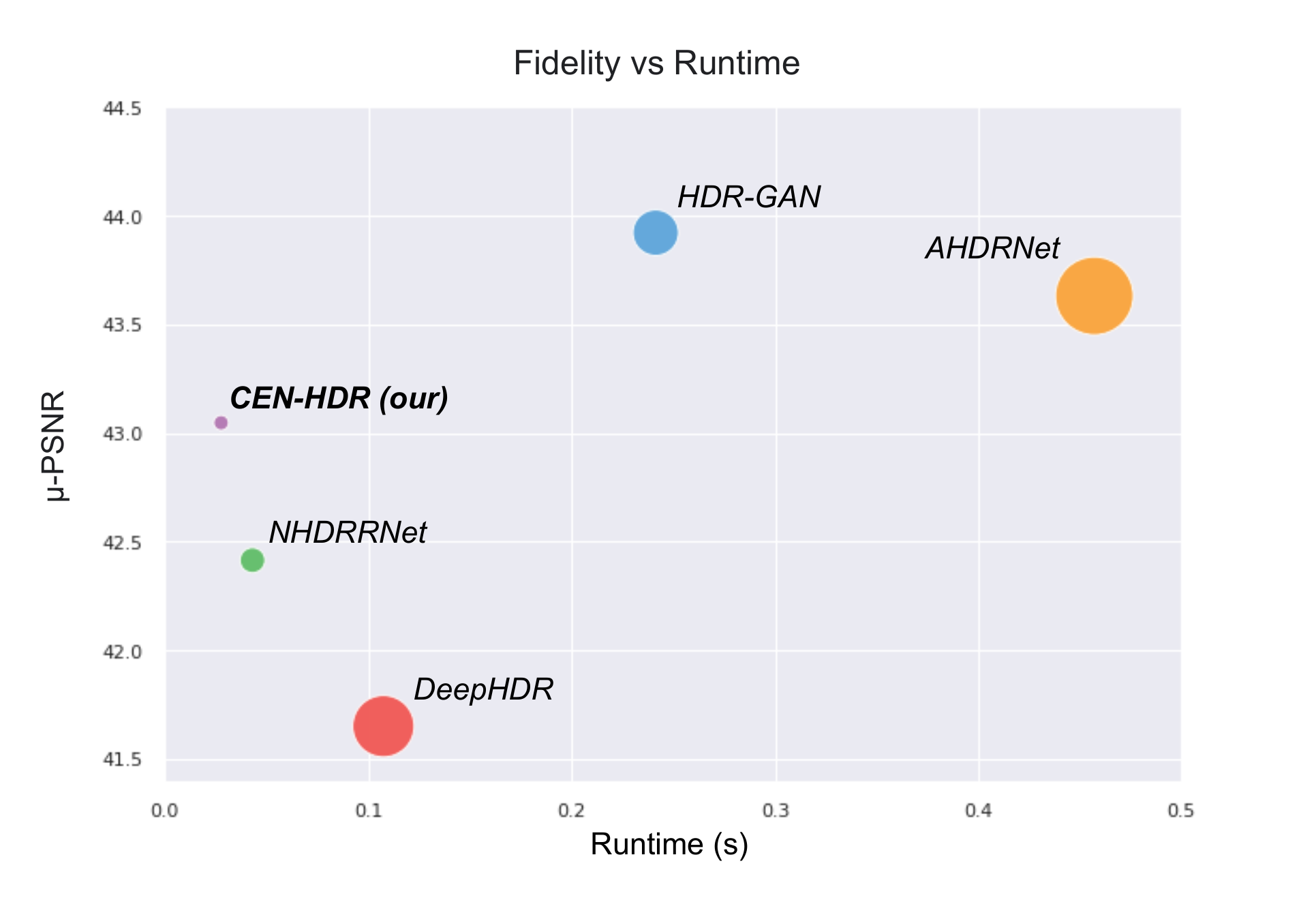}
\caption{Comparison of the proposed CEN-HDR solution with other HDR merging methods. The X-axis represents the mean runtime using the M1 NPU with an input size of 1280x720 pixels. The Y-axis is the fidelity score on the test images from \cite{Kalantari2017} dataset. The best solutions tend to be in the upper left corner. The radius of circles represents the number of operations, the smaller the better. }
\label{fig:fidelity_vs_runtime}
\end{figure}

\setlength{\tabcolsep}{4pt}
\begin{table}[h]
\begin{center}
\caption{Comparison of the proposed CEN-HDR architecture performances with and without knowledge distillation. In the first case, the network is trained with the HDR ground-truth proposed in \cite{Kalantari2017}. In the second case, we also use the prediction of HDR-GAN \cite{hdrgan} as label (Eq. \ref{eq:lossfunction}). In both cases, the network is trained for 500 epochs using the $l_1$ criterion.}
\label{table:kd}
\scalebox{0.9}{
\begin{tabular}{l|lllll}
\hline\noalign{\smallskip}
Training method & $\mu$-PSNR & PSNR & $\mu$-SSIM  & SSIM & HDR-VDP2 \\
\noalign{\smallskip}
\hline
\noalign{\smallskip}
w/o knowledge distillation & 40.8983 &  40.0298  & 0.9772 & 0.9926 & 62.17 \\
with knowledge distillation  & 43.0470 & 40.5335  & 0.9908 & 0.9956  & 64.34  \\ 

\hline
\end{tabular}}
\end{center}
\end{table}
\setlength{\tabcolsep}{1.4pt}

\section{Experimental Results}

\setlength{\tabcolsep}{4pt}

\begin{table}[h]
\begin{center}
\caption{Quantitative comparison with lightweight state-of-the-art methods on the Kalantari2017\cite{Kalantari2017} test samples. PSNR and SSIM are calculated in the linear domain while $\mu$-PSNR and $\mu$-SSIM are calculated after $\mu$-law tone mapping (Eq.~\ref{eq:mulaw}). For compared methods, the results are from \cite{hdrgan}. PU-PSNR and PU-SSIM are calculated applying the encoding function proposed in \cite{pu21}.}
\label{table:quant}
\scalebox{0.9}{
\begin{tabular}{l|rrrrrrr}
\hline\noalign{\smallskip}
Method & $\mu$-PSNR & PU-PSNR & PSNR& $\mu$-SSIM & PU-SSIM & SSIM & HDR-VDP2\\
\noalign{\smallskip}
\hline
\noalign{\smallskip}
Sen et al.\cite{PatchBasedHDR} & 40.80 & 32.47 & 38.11 & 0.9808 &0.9775 &  0.9721 &  59.38 \\
Hu et al.\cite{Hu} & 35.79 & $-$ & 30.76 & 0.9717 & $-$ & 0.9503 & 57.05 \\
Kalantari et al.\cite{Kalantari2017} & 42.67 & 33.82 & 41.23 & 0.9888 & 0.9832 & 0.9846  & 65.05  \\
DeepHDR\cite{deephdr} & 41.65 & 31.36 & 40.88  & 0.9860 & 0.9815 & 0.9858   & 64.90\\
NHDRRNet\cite{nhdrrnet} & 42.41 & $-$ & $-$ & 0.9887 & $-$ & $-$ &  61.21\\
AHDRNet\cite{ahdrnet}  & 43.61 & 33.94 & 41.03 & 0.9900  & 0.9855 & 0.9702  & 64.61 \\
HDRGAN\cite{hdrgan} & 43.92 & 34.04 & 41.57  & 0.9905 & 0.9851 & 0.9865  & 65.45 \\
CEN-HDR(our) & 43.05 & 33.23 & 40.53  & 0.9908 & 0.9821 & 0.9856  &  64.34\\
\hline
\end{tabular}}
\end{center}
\end{table}
\setlength{\tabcolsep}{1.4pt}

\subsection{Fidelity performance}
In Table~\ref{table:quant} and Fig.~\ref{fig:qualitative_comparison}, the proposed CEN-HDR solution is compared against seven lightweight state-of-the-art methods: \cite{PatchBasedHDR} and \cite{Hu} are based on input patch registration methods. \cite{Kalantari2017} is based on a sequential CNN, the inputs need first to be aligned thanks to an optical flow algorithm. For \cite{deephdr}, the background of each LDR input is aligned by homography before being fed to an encoder-decoder-based CNN. \cite{nhdrrnet} proposes an encoder-decoder architecture with a non-local attention module. \cite{ahdrnet} is a CNN based on an attention block for features registration and on multiple dilated residual dense blocks for merging.  \cite{hdrgan} is the first GAN-based approach for HDR merging with a deep supervised HDR method. Quantitative evaluation is done using objective metrics. The standard peak signal-to-noise ratio (PSNR) and structural similarity (SSIM) are computed both directly in the linear domain and after tone mapping by applying the $\mu$-law function (Eq.~\ref{eq:mulaw}). The HDR-VDP2\cite{hdrvdp2} metric predicts the quality degradation with the respect to the reference image. We set the diagonal display size to 24 inches and the viewing distance to 0.5 meter. In addition, PU-PSNR and PU-SSIM are calculated by applying the encoding function proposed in \cite{pu21} with a peak luminance set to 4000. 

\begin{figure}[ht]
\centering
\includegraphics[height=10.6cm]{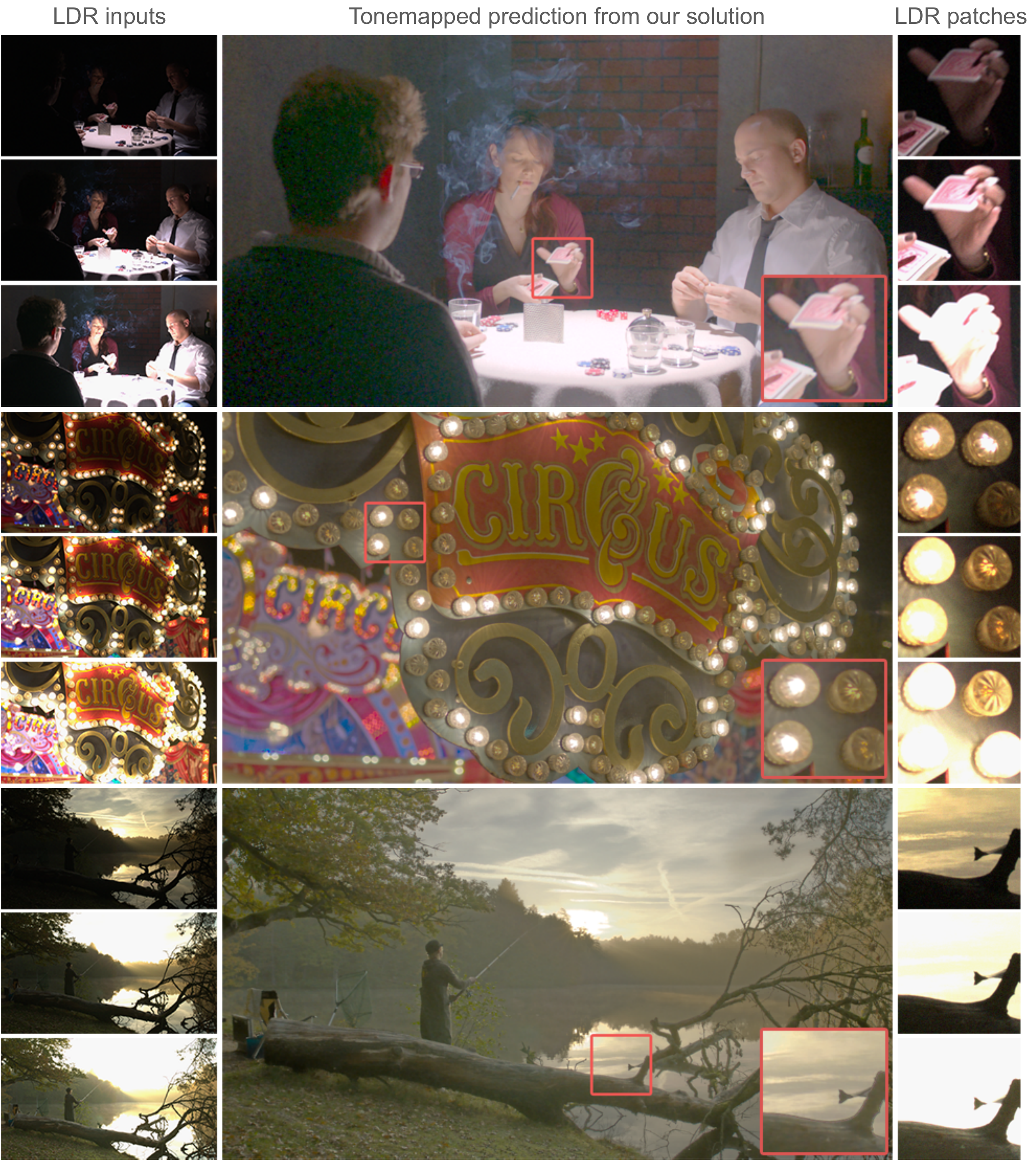}
\caption{Qualitative results of the proposed CEN-HDR solution on samples from the NTIRE\cite{ntire22} challenge dataset. The ground truth images are not provided. We notice that the high sensor noise present in the input images is not fully corrected in the dark areas of the output HDR images. Moreover, the motion blur introduced in the dataset produces less sharp characteristics in the output image.}
\label{fig:qualitative_ntire}
\end{figure}

In Fig.~\ref{fig:qualitative_ntire}, we present the results for 3 test scenes of the NTIRE\cite{ntire22} dataset. While the network can produce images with a high dynamic range, we notice that the high sensor noise present in the input images is not fully corrected in the dark areas of the output HDR images. Moreover, the motion blur introduced in the dataset produces less sharp characteristics in the output image.

In Table~\ref{table:attmoduleabltion} we study the effect of the attention module on the performance of proposed HDR deghosting architecture. SCRAM-C and SCRAM-S respectively correspond to the SCRAM module composed only of the channel attention branch and the spatial attention branch. The proposed SCRAM allows to achieve a similar quality as the spatial attention module proposed in AHDRNet[28], while Table 5 shows that the inference cost of SCRAM is lower.

\subsection{Efficiency comparison}
As we want to propose an efficient HDR generation method, in Table~\ref{table:time} we compare the computation cost and the inference time of our network with state-of-the-art HDR networks that achieve similar performance on quality metrics.

The number of operations and parameters are measured using the script provided by the NTIRE\cite{ntire22} challenge. To evaluate runtimes, all the compared networks are executed on the Neural Processing Unit (NPU) of a MacBook Pro (2021) powered with an M1 chip. The time shown is the average for 500 inference runs after a warm-up of 50 runs. The input size is set to $1280 \times 720$ pixels. The gamma-projection of LDR inputs  (Eq.~\ref{eq:hi}) and the tone mapping of the HDR output are included in the inference time measurement. Note that the background alignment of inputs frame using homography for \cite{deephdr} is not included.

\setlength{\tabcolsep}{4pt}
\begin{table}[h]
\begin{center}
\caption{Inference cost comparison of the proposed CEN-HDR solution against  state-of-the-art lightweight deep learning-based methods. The number of operations and parameters are measured using the script provided by the NTIRE\cite{ntire22} challenge. To measure the inference time, all the compared networks are executed on an M1 NPU. The input size is set to $1280 \times 720$ pixels.}
\label{table:time}
\begin{tabular}{l|rrrr}
\hline\noalign{\smallskip}
Method & Num. of params. & Num. of op. (GMAccs) & Runtime(s) & FPS\\
\noalign{\smallskip}
\hline
\noalign{\smallskip}
DeepHDR\cite{deephdr} & 14618755 & 843.16  & 0.1075 & 9.30\\
AHDRNet\cite{ahdrnet}  & 1441283 & 1334.95  & 0.4571 & 2.18\\
NHDRRNet\cite{nhdrrnet} & 7672649 & 166.11 & 0.0431 & 23.20\\
HDR-GAN\cite{hdrgan} & 2631011 & 479.78 & 0.2414 & 4.14 \\
CEN-HDR(our) & 282883 & 78.36 & 0.0277 & 36.38\\
\hline
\end{tabular}
\end{center}
\end{table}
\setlength{\tabcolsep}{1.4pt}

 Table \ref{table:ntirearch} compares the number of parameters and operations of the proposed CEN-HDR solution with recent efficient methods \cite{Li_2022_CVPR,Yan_2022_CVPR,progresiveNTIRE}. The input size is set to $1900 \times 1060$ pixels corresponding to the size of the inputs from the dataset proposed by \cite{ntire22}.
 For compared methods \cite{Li_2022_CVPR,Yan_2022_CVPR,progresiveNTIRE} the measurements are provided by the NTIRE\cite{ntire22} challenge. We could not compare the inference time of the CEN-HDR solution with these three architectures as they were recently proposed and their implementation is not yet available.
 
Fig.~\ref{fig:fidelity_vs_runtime} compares the trade-off between fidelity to the ground truth label and runtime of the proposed CEN-HDR solution with other HDR merging methods. The X-axis represents the mean runtime using an M1 NPU with an input size of 1280x720 pixels. The Y-axis is the fidelity score on the test images from \cite{Kalantari2017} dataset. The best solutions tend to be in the upper left corner. The radius of circles represents the number of operations. Our solution is shown as the best solution for real-time HDR merging with a high-fidelity score.

\setlength{\tabcolsep}{4pt}
\begin{table}[h]
\begin{center}
\caption{Inference cost comparison of the proposed CEN-HDR solution versus recent efficient merging networks. The number of operations and parameters for \cite{Li_2022_CVPR,Yan_2022_CVPR,progresiveNTIRE} and our solution are computed following the method described in \cite{ntire22}.The input size is set to $1900 \times 1060$ pixels.}
\label{table:ntirearch}
\begin{tabular}{l|rrrr}
\hline\noalign{\smallskip}
Method & Num. of params. & Num. of op. (GMAccs) \\
\noalign{\smallskip}
\hline
\noalign{\smallskip}
GSANet\cite{Li_2022_CVPR} & 80650 & 199.39   \\
Yan et al.\cite{Yan_2022_CVPR}  & 18899000 & 156.12  \\
Yu et al.\cite{progresiveNTIRE} & 1013250 & 199.88   \\
CEN-HDR(our) & 282883 & 128.78 \\
\hline
\end{tabular}
\end{center}
\end{table}
\setlength{\tabcolsep}{1.4pt}

\setlength{\tabcolsep}{4pt}
\begin{table}[h]
\begin{center}
\caption{Inference cost comparison of attention modules. Spatial and Channel attention modules are studied by feeding a tensor of size $(1, \frac{H}{4}, \frac{W}{4})$  corresponding to the concatenation of the reference and non-reference tensors after the encoding step.}
\label{table:attentionmodulecmp}
\scalebox{0.9}{
\begin{tabular}{l|rr|rrr}
\hline\noalign{\smallskip}
Method & \multicolumn{2}{c|}{Attention type} & params. & GMAccs & Runtime(s)\\\cline{2-3}
 & Spatial & Channel &  &  & \\
\noalign{\smallskip}
\hline
\noalign{\smallskip}
AHDRNet attention \cite{ahdrnet} & \checkmark & & 55392  & 20.772 & 0.0085\\
EPSANet\cite{PSA} & \checkmark & & 42560 & 15.768 & 0.0111\\
SK attention \cite{skattention} & & \checkmark  & 125984  & 43.104 & 0.0155\\
Double attention\cite{A2Attention} & \checkmark & \checkmark & 33216 & 12.456 & 0.0101\\
CBAM\cite{CBAM}  & \checkmark & \checkmark  & 22689  & 7.525 & 0.0734\\
BAM\cite{BAM} & \checkmark  & \checkmark & 17348 & 5.008 & 0.0060\\

\hline
\end{tabular}}
\end{center}
\end{table}
\setlength{\tabcolsep}{1.4pt}

\setlength{\tabcolsep}{4pt}
\begin{table}[h]
\begin{center}
\caption{Effect of the attention module on the performance of proposed HDR deghosting network. SCRAM-C and SCRAM-S respectively correspond to the SCRAM module composed only of the channel attention branch and the spatial attention branch.}
\label{table:attmoduleabltion}
\scalebox{0.9}{
\begin{tabular}{l|rrrr}
\hline\noalign{\smallskip}
Method & $\mu$-PSNR & PSNR& $\mu$-SSIM  & SSIM \\
\noalign{\smallskip}
\hline
\noalign{\smallskip}
Without attention module & 42.12  & 39.95 &  0.9850 & 0.9823 \\
AHDRNet\cite{ahdrnet} attention module & 42.94 & 40.49  & 0.9903 & 0.9852 \\
SCRAM-C  & 42.32 & 40.14 & 0.9854 & 0.9829 \\
SCRAM-S  & 42.89 & 40.41 & 0.9884 & 0.9835 \\
SCRAM  & 43.05 & 40.53  & 0.9908 & 0.9856 \\

\hline
\end{tabular}}
\end{center}
\end{table}
\setlength{\tabcolsep}{1.4pt}

\section{Conclusions}

In this paper, we propose CEN-HDR, a novel computationally efficient HDR merging network able to correct the ghost effect caused by large object motions in the scene and camera motion. 
The proposed lightweight network architecture effectively succeeds in generating real-time HDR images with a dynamic range close to that of the original scene.
By integrating the knowledge distillation methods in our training scheme, we demonstrate that the majority of the representation capabilities of a large HDR merging network can be transferred into a lighter network, opening the door to real-time HDR embedded systems.

\clearpage
%
%
\bibliographystyle{splncs04}
\bibliography{bib}
\end{document}